# Energy-efficient Machine Learning in Silicon: A Communications-inspired Approach

**Naresh Shanbhag**  SHANBHAG@ILLINOIS.EDU
University of Illinois at Urbana-Champaign, Urbana, IL 61801 USA

## Abstract

This position paper advocates a communications-inspired approach to the design of machine learning systems on energy-constrained embedded 'always-on' platforms. The communications-inspired approach has two versions - 1) a deterministic version where existing low-power communication IC design methods are repurposed, and 2) a stochastic version referred to as Shannon-inspired *statistical information processing* employing information-based metrics, *statistical error compensation* (SEC), and retraining-based methods to implement ML systems on stochastic circuit/device fabrics operating at the limits of energy-efficiency. The communications-inspired approach has the potential to fully leverage the opportunities afforded by ML algorithms and applications in order to address the challenges inherent in their deployment on energy-constrained platforms.

## 1. Introduction

Machine learning (ML)-based systems are transforming the way we live and interact with the world around us. In many tasks, such as those in computer vision, machines have begun to exceed human performance (Silver et al., 2016). However, machines have much catching up to do when energy costs are accounted for. While it is difficult to accurately estimate the energy cost of the AlphaGo system developed by Google DeepMind when it beat the human champion recently in the ancient game of Go, one can safely assume that the machine consumed about four-orders-of-magnitude higher power (1202 CPUs and 176 GPUs (Silver et al., 2016)) as compared to the nominally quoted power of 20 W for the human brain. If ML systems need to become pervasive in our lives then it is imperative that this energy cost be significantly reduced. The availability of such low-energy realizations of ML systems will enable its deployment on embedded platforms such as biomedical devices, wearables, autonomous vehicles, IoT and many others. Not surprisingly, a number of integrated circuit (IC) implementations of ML kernels and algorithms have appeared recently (Chen et al., 2016; Kaul et al., 2016; Park et al., 2016) that have set energy-efficiency records. However, much work still remains to be done as the energy gap between these realizations and that achieved by the human brain remains huge. In particular, the search for minimum energy realizations of ML systems needs to be done systematically. The low-energy ML design space is complex as it encompasses deeply intertwined issues at the algorithmic, architectural, circuit and the device level. The mainstream approach today is to treat the problem of energy-efficient ML implementation as yet another problem in energy-efficient computing. We believe that there are substantial gains to be made if This work was supported in part by Systems on Nanoscale Information fabriCs (SONIC), one of the six SRC STARnet Centers, sponsored by MARCO and DARPA.

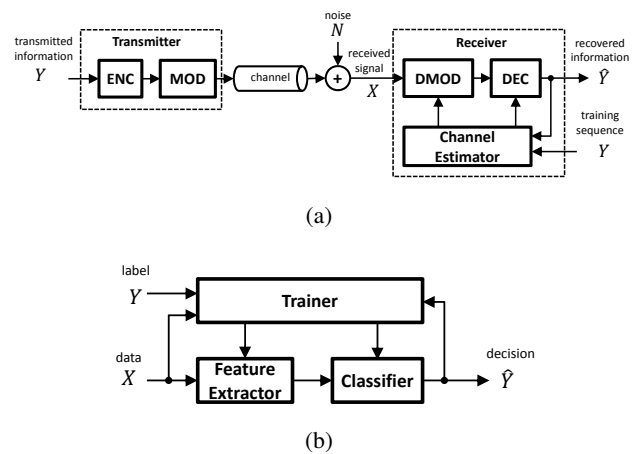

*Figure 1.* Viewing a communication receiver as an inference system: (a) the communication link, and (b) a ML system.

one were to repurpose the vast body of knowledge accumulated over two and a half decades by the designers of low-power communication and signal processing systems and ICs (A.P. Chandrakasan & Brodersen, 1992; Shanbhag, 1998; Parhi, 1999). This position paper makes the case for employing a communications-inspired approach in order to explore the design of energy-efficient ML in nanoscale silicon CMOS and emerging beyond CMOS device fabrics.

The communications-inspired approach in based on drawing parallels between a communication receiver and an inference kernel as shown in Fig. 1. A communication receiver infers the transmitted symbols $Y$ from the received signal $X$, much as a ML system infers the class label $Y$ from the observed data $X$. In both systems, the process of inference needs to be accomplished in the presence of random noise and incomplete data. Both systems need an element of learning/training to be present in order to incorporate time-varying/unknown data statistics/model into the decision making process. Communication receivers commonly employ statistical estimation procedures to learn the channel parameters, which are then employed for data recovery. Furthermore, the stochastic gradient descent (SGD) (Mathews & Xie, 1993; Keuper & Pfreundt, 2015) is commonly employed in both systems due to its ease of implementation and robustness. There is one key difference between the two systems though. In communication systems, the data $X$'s statistics can be *engineered* via proper coding and modulation in the transmitter. This allows such receiver to operate with well-structured signal, channel and noise models, which lowers its complexity and energy consumption, while enhancing its accuracy. This flexibility may not be present in general ML scenarios. Nevertheless, the similarities between the two are substantial enough to warrant a closer look at low-power communication receiver design techniques and see which ones might be repurposed for ML systems.

In the discussion above, one assumes a deterministic circuit fabric. Recent IC implementations (Chen et al., 2016; Kaul et al., 2016; Park et al., 2016) do in fact fit this model. However, this assumption can be relaxed in case of ML systems due to their inherent ability to operate in the presence of incomplete or noisy data. This ability can be leveraged to address the statistical behavior of circuit/device fabrics that arises when these operate at the limits of energy efficiency. Such ultimate low-energy fabrics is referred to as *stochastic fabrics* or *low-SNR circuit fabrics*. Indeed, statistical behavior in such fabrics can arise when:

- operating at very low voltages (Dreslinski et al., 2010) or low area (Roy et al., 2013), both of which result in computational errors, and/or
- designing systems with emerging devices (Roy et al., 2013; Wei et al., 2013) which tend to be intrinsically statistical in nature due to nanoscale imperfections such as variations and defects, and/or
- embedding computation into memory (in-memory computing (Kang et al., 2014)) and sensing (in-sensor computing (Hu et al., 2012)) substrates in order to drastically reduce/eliminate data movement.

We refer to such ultimate low-energy fabrics as *stochastic fabrics*. The statistical behavior of stochastic fabrics needs to be compensated for much as a communication receiver compensates for the statistical behavior of the channel. The communications-inspired view opens up the possibility of taking the connections between ML and communications to another level by treating the circuit fabric itself as a noisy channel on which to extract information from data. We refer to this second approach as Shannon-inspired *statistical information processing* (Shanbhag et al., 2010). Statistical information processing involves the use of information-based metrics, *statistical error compensation* (SEC) (Hegde & Shanbhag, 2001), and retraining approaches such as data-driven hardware resiliency (DDHR) (Wang et al., 2015) to enhance robustness. One intellectually satisfying aspect of statistical information processing is the potential for developing a comprehensive foundation for reliable information processing on stochastic fabrics much as Claude Shannon (Shannon, 1948) established one for reliable communications over a noisy channel. Such a foundation needs to provide fundamental bounds on the information processing capacity, energy-efficiency, robustness, as well as practical design techniques, e.g., SEC and DDHR, to approach these bounds.

This paper advocates a communications-inspired approach to the design of energy-efficient ML systems on both deterministic and stochastic fabrics. Doing so will bring together methodologies such as low-power signal processing algorithms and architectures (Parhi, 1999), algorithm transforms (Shanbhag, 1998), low-power integrated circuit (IC) design (A.P. Chandrakasan & Brodersen, 1992), information-based design metrics, *statistical error compensation* (SEC) and others to systematically explore the design space in order to determine minimum energy realizations.

## 2. Machine Learning on Deterministic Fabrics

The design of communication receiver ICs begins with algorithm design employing statistical signal processing techniques such as estimation and detection to meet a specific system design metric such as the bit-error rate (BER) $p_e = P\{Y \neq \hat{Y}\}$ (see Fig. 1). The use of an information-based metric (BER) and its intrinsically statistical nature makes it possible to reduce algorithmic com-plexity right from the start. Redundant algorithmic operations are eliminated or substituted with approximate ones so as to leave the BER unaltered. Machine learning systems employ an accuracy metric $p_{det}$ the probability of detection, and therefore can benefit from such approximations. Indeed, "approximate computing" (Venkataramani et al., 2015) strives to build a methodology to systematize and repurpose these concepts which are well-known and well-practiced for decades by communication IC designers. The result of this step is a floating-point algorithm meeting the system requirements on BER and other metrics.

Next, *fixed-point analysis* is employed to minimize the precision of computation and storage. Indeed, minimizing precision (Gupta et al., 2015) is an effective approach to reduce energy. The goal of this step is to minimize the BER difference between the floating-point and a fixed-point algorithm. Precisions is typically obtained via trial-and-error. Insights on what algorithmic aspects determine the precision tend to be lost in this process. However, for communications and ML algorithms, it is possible to obtain analytical bounds on precision. For example, the bounds on the precision $B_{WUD}$ of the weight-update unit of the popular least mean-squared (LMS) algorithm (Goel & Shanbhag, 1998) is given by:

$$B_{WUD} \geq \frac{1}{2}log_2\left(\frac{1}{\mu^2\sigma_y^2\sigma_x^2}\right) + \frac{SNR_{fl}(dB)}{6} \quad (1a)$$

where $\mu$ is the step-size, $\sigma_x^2$ and $\sigma_y^2$ are variances of the input $X$ and desired signal $Y$, respectively, and $SNR_{fl}(dB)$ is the SNR of the floating point algorithm in dBs. Minimum precision requirements are thus obtained without resorting to expensive simulations. In a similar fashion, it is possible to obtain bounds for other SGD-based on-line learning algorithms.

The fixed-point algorithm can be described using a data flow-graph (DFG) or a control and data flow graph (CDFG). An almost infinite variety of architectures can be systematically obtained from a DFG using algorithm transforms (Parhi, 1999) such as unfolding, folding, pipelining, systolization, among others. ML algorithms tend to have a regular DFG (see Fig. 2). This opens up the possibility of realizing *systolic architectures* (Kung, 1982) for many ML algorithms. Some work already exists (Jones et al., 1994; Kung & Hwang, 1989). Systolic architectures are regular, have local interconnections, and can be designed to minimize data movement. The process of mapping a regular DFG to a systolic architectures involves the selection of a *processor vector* **p**, the *iteration vector* **d** and the *schedule vector* **s**, satifying the constraints $\mathbf{p}^T\mathbf{d} = 0$, $\mathbf{s}^T\mathbf{d} \neq 0$, and implying that the DFG node **v** is mapped to processor $\mathbf{p}^T\mathbf{v}$ in the cycle $\mathbf{s}^T\mathbf{v}$. Indeed, one can derive the recently proposed architectures (Chen et al., 2016; Murmann et al., 2015) by formulating the DFG of a convolutional neural network (CNN) (LeCun et al., 1998) (see Fig. 2(b)), and assigning appropriate values to **p**, **d**, and **s**, along with the folding transform. These design methodologies for communication ICs can be repurposed for the design of energy-efficient ML systems in silicon.

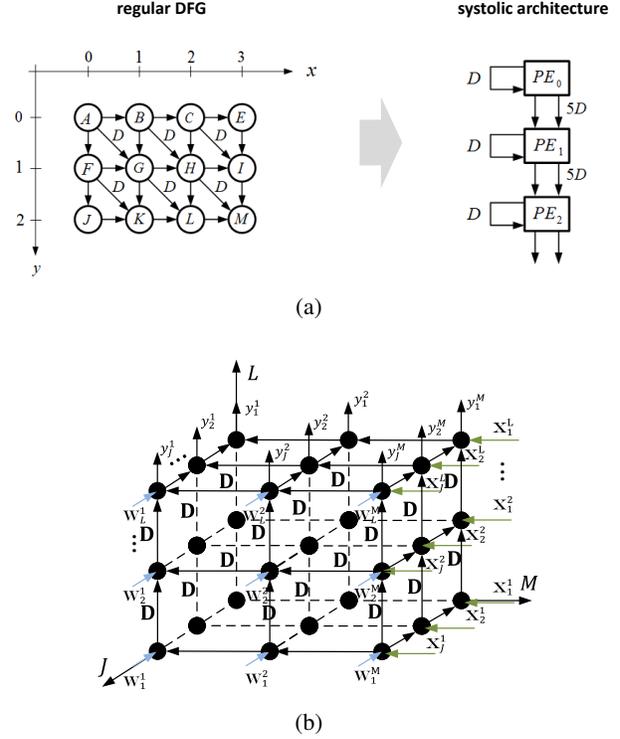

Figure 2. Systolization: (a) a regular DFG mapped to a systolic architecture via specific choices of vectors $\mathbf{p} = [1\ 0]^T$, $\mathbf{d} = [0\ 1]^T$ and $\mathbf{s} = [1\ 0]^T$, where $D$ is a 1-sample delay element, and (b) the DFG of the C-layer of a CNN with each node being a dot-product computation.

## 3. Machine Learning on Stochastic Fabrics

The communications-inspired approach presents a unique opportunity when implementing ML on deeply scaled nanofabrics that operate at the limits of energy efficiency where a transition into non-determinism occurs. For example, near/subthreshold voltage (Dreslinski et al., 2010) operation in CMOS results approximately $10\times$ reduction in energy but at the expense of up to $20\times$ increase in delay variations. This variability eventually translates into observable errors in computation, storage, and communications. We refer to such circuit and device substrate as *stochastic fabrics*, and the errors themselves as *fabric noise*. ML algorithms' intrinsic robustness to data noise enables it to absorb the impact of fabric noise. This feature, referred to popularly as 'error-tolerance', can be exploited to some extent by approaches such as approximate computing (Venkataramani et al., 2015) as well. However, it is possible to reduce the energy consumption even further by operating the circuit fabric at a point where the intrinsic error-tolerance of the algorithm is exceeded. At this point, corrective measures, i.e., error compensation methods, need to be incorporated. Conventional fault-tolerance techniques such as $N$-modular redundancy are ineffective as these have a high energy-cost, and do not account for the unique attributes of ML algorithms. A Shannon-inspired approach to error compensation turns out to be most effective.

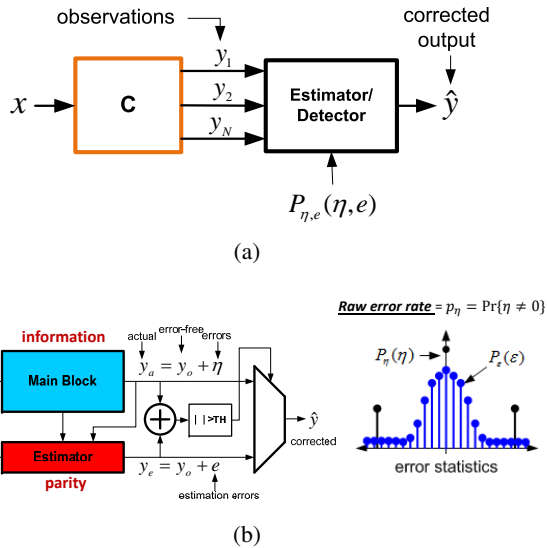

Figure 3. Shannon-inspired statistical error compensation (SEC): (a) a general framework, and (b) algorithmic noise-tolerance (ANT).

In the past, we have proposed the notion of treating the stochastic circuit fabric as a noisy communication channel (Shanbhag, 1996) and develop Shannon-inspired statistical error compensation (SEC) techniques (see Fig. 3(a))(Hegde & Shanbhag, 2001; Shim et al., 2004; Varatkar et al., 2010) to compensate for the resulting errors at the algorithmic and architectural levels. Prototype ICs (see Fig. 4) demonstrating these ideas have been implemented. These demonstrate that computational error rates, defined as the probability of an incorrect output, of 60% (Abdallah & Shanbhag, 2013) and in specific cases (see Fig. 4(b)), up to 80% (Kim et al., 2015) can be compensated for by applying techniques based on statistical estimation and detection. SEC techniques have shown to result in energy savings ranging from $3\times$-to-$6\times$ over designs that work on deterministic fabrics.

The ability to compensate for such high computational error rates motivates the idea of *in-situ data analytics*, where computation is deeply embedded into the same substrate where data is stored or being acquired, e.g., *in-memory* (Kang et al., 2014) and *in-sensor* computing (Hu et al., 2012). Such subtrates are not particularly well-suited for deterministic von Neumann style computing but fits the Shannon-inspired style. Thus, SEC leverages Shannon theory to develop techniques to compensate for errors that cannot be absorbed by the intrinsic error-tolerance of the algorithm. This key aspect distinguishes it from techniques that seek to work within the error-tolerance envelope of the algorithm. SEC techniques can be made adaptive in order to track variations in the data and error statistics. ML-based SEC techniques can also be developed.

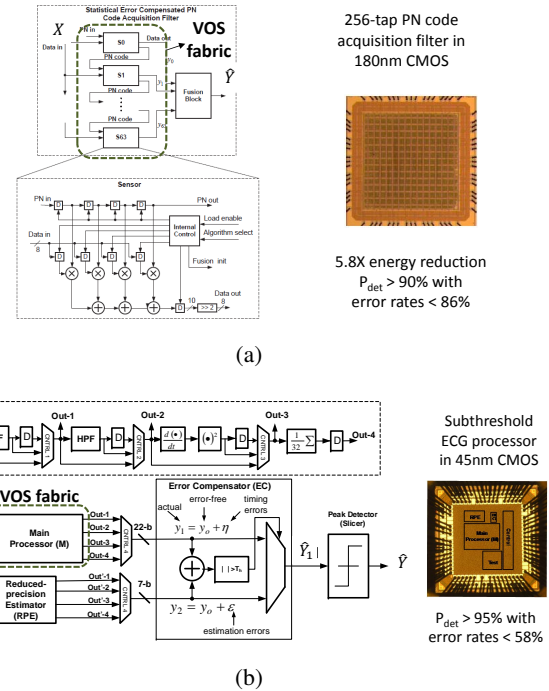

Figure 4. Statistical error compensation (SEC) based IC prototypes: (a) 256-tap PN code acquisition in $180\,\text{nm}$ CMOS, and (b) ECG processor in $45\,\text{nm}$ CMOS.

Another approach is DDHR (Wang et al., 2015) that employs retraining to obtain parameters of the algorithm to compensate for both data and fabric noise. Both SEC and DDHR leverage the statistical nature of system and application metrics, and may even be combined in a synergistic fashion.

## 4. Summary

ML systems have unique properties that it shares with communication systems. There is much to be gained by exploiting the connections between the two when exploring energy efficient on-device implementations of ML systems.